\pdfoutput=1
\documentclass[11pt,letterpaper]{article}
\usepackage{emnlp2017}
\usepackage{times}
\usepackage{latexsym}
\usepackage{tipa}
\usepackage{amsmath}
\usepackage[draft]{todonotes} 
\usepackage{booktabs}

\emnlpfinalcopy

\def\emnlppaperid{***}

\title{Massively Multilingual Neural Grapheme-to-Phoneme Conversion}

\author{Ben Peters \\ Saarland University \\ Saarbr\"ucken, Germany \\ \tt{benzurdopeters@gmail.com}
		\And Jon Dehdari \and Josef van Genabith \\ DFKI \& Saarland University \\ Saarbr\"ucken, Germany \\ \tt{firstname.lastname@dfki.de}
        }

\date{}

\begin{document}

\maketitle

\begin{abstract}
Grapheme-to-phoneme conversion (g2p) is necessary for text-to-speech and automatic speech recognition systems. Most g2p systems are monolingual: they require language-specific data or handcrafting of rules. Such systems are difficult to extend to low resource languages, for which data and handcrafted rules are not available. As an alternative, we present a neural sequence-to-sequence approach to g2p which is trained on spelling--pronunciation pairs in hundreds of languages. The system shares a single encoder and decoder across all languages, allowing it to utilize the intrinsic similarities between different writing systems. We show an 11\% improvement in phoneme error rate over an approach based on adapting high-resource monolingual g2p models to low-resource languages. Our model is also much more compact relative to previous approaches.
\end{abstract}

\section{Introduction}
Accurate grapheme-to-phoneme conversion (g2p) is important for any application that depends on the sometimes inconsistent relationship between spoken and written language. Most prominently, this includes text-to-speech and automatic speech recognition. Most work on g2p has focused on a few languages for which extensive pronunciation data is available \citep[\it inter alia]{bisani2008joint,novak2016phonetisaurus,rao2015grapheme,yao2015sequence}. Most languages lack these resources. However, a low resource language's writing system is likely to be similar to the writing systems of languages that do have sufficient pronunciation data. Therefore g2p may be possible for low resource languages if this high resource data can be properly utilized. 

We attempt to leverage high resource data by treating g2p as a multisource neural machine translation (NMT) problem. The source sequences for our system are words in the standard orthography in any language. The target sequences are the corresponding representation in the International Phonetic Alphabet (IPA). Our results show that the parameters learned by the shared encoder--decoder are able to exploit the orthographic and phonemic similarities between the various languages in our data.

\section{Related Work}
\subsection{Low Resource g2p}
Our approach is similar in goal to \newcite{deri2016grapheme}'s model for adapting high resource g2p models for low resource languages. They trained weighted finite state transducer (wFST) models on a variety of high resource languages, then transferred those models to low resource languages, using a language distance metric to choose which high resource models to use and a phoneme distance metric to map the high resource language's phonemes to the low resource language's phoneme inventory. These distance metrics are computed based on data from Phoible \cite{phoible} and URIEL \cite{littell2017uriel}.

Other low resource g2p systems have used a strategy of combining multiple models. \newcite{schlippe2014combining} trained several data-driven g2p systems on varying quantities of monolingual data and combined their outputs with a phoneme-level voting scheme. This led to improvements over the best-performing single system for small quantities of data in some languages. \newcite{jyothilow} trained recurrent neural networks for small data sets and found that a version of their system that combined the neural network output with the output of the wFST-based Phonetisaurus system \cite{novak2016phonetisaurus} did better than either system alone.

A different approach came from \newcite{kim2012universal}, who used supervised learning with an undirected graphical model to induce the grapheme--phoneme mappings for languages written in the Latin alphabet. Given a short text in a language, the model predicts the language's orthographic rules. To create phonemic context features from the short text, the model na\"ively maps graphemes to IPA symbols written with the same character, and uses the features of these symbols to learn an approximation of the phonotactic constraints of the language. In their experiments, these phonotactic features proved to be more valuable than geographical and genetic features drawn from WALS \cite{wals}.

\subsection{Multilingual Neural NLP}
In recent years, neural networks have emerged as a common way to use data from several languages in a single system. Google's zero-shot neural machine translation system \cite{johnson2016google} shares an encoder and decoder across all language pairs. In order to facilitate this multi-way translation, they prepend an artificial token to the beginning of each source sentence at both training and translation time. The token identifies what language the sentence should be translated to. This approach has three benefits: it is far more efficient than building a separate model for each language pair; it allows for translation between languages that share no parallel data; and it improves results on low-resource languages by allowing them to implicitly share parameters with high-resource languages. Our g2p system is inspired by this approach, although it differs in that there is only one target ``language'', IPA, and the artificial tokens identify the language of the source instead of the language of the target.

Other work has also made use of multilingually-trained neural networks. Phoneme-level polyglot language models \cite{tsvetkov2016polyglot} train a single model on multiple languages and additionally condition on externally constructed typological data about the language. \newcite{ostling2017continuous} used a similar approach, in which a character-level neural language model is trained on a massively multilingual corpus. A language embedding vector is concatenated to the input at each time step. The language embeddings their system learned correlate closely to the genetic relationships between languages. However, neither of these models was applied to g2p.

\section{Grapheme-to-Phoneme}
g2p is the problem of converting the orthographic representation of a word into a phonemic representation. A phoneme is an abstract unit of sound which may have different realizations in different contexts. For example, the English phoneme \textipa{/p/} has two phonetic realizations (or allophones):

\begin{itemize}
\item \textipa{[p\super h]}, as in the word `pain' \textipa{[p\super h eI n]}
\item \textipa{[p]}, as in the word `Spain' \textipa{[s p eI n]}
\end{itemize}

English speakers without linguistic training often struggle to perceive any difference between these sounds. Writing systems usually do not distinguish between allophones: \textipa{[p\super h]} and \textipa{[p]} are both written as $\langle$p$\rangle$ in English. The sounds are written differently in languages where they contrast, such as Hindi and Eastern Armenian.

Most writing systems in use today are glottographic, meaning that their symbols encode solely phonological information\footnote{The Chinese script, in which characters have both phonological form and semantic meaning, is the best-known exception.}. But despite being glottographic, in few writing systems do graphemes correspond one-to-one with phonemes. There are cases in which multiple graphemes represent a single phoneme, as in the word \emph{the} in English:

\begin{center}
\begin{tabular}{cc}
th \ & e \\
\textipa{D} & \textipa{@}
\end{tabular}
\end{center}

There are cases in which a single grapheme represents multiple phonemes, such as syllabaries, in which each symbol represents a syllable.

In many languages, there are silent letters, as in the word \emph{hora} in Spanish: 

\begin{center}
\begin{tabular}{cccc}
h & o & r & a \\
- & \textipa{o} & \textipa{R} & \textipa{a}
\end{tabular}
\end{center}

There are more complicated correspondences, such as the silent \textit{e} in English that affects the pronunciation of the previous vowel, as seen in the pair of words \emph{cape} and \emph{cap}. 

It is possible for an orthographic system to have any or all of the above phenomena while remaining unambiguous. However, some orthographic systems contain ambiguities. English is well-known for its spelling ambiguities. Abjads, used for Arabic and Hebrew, do not give full representation to vowels.

Consequently, g2p is harder than simply replacing each grapheme symbol with a corresponding phoneme symbol. It is the problem of replacing a grapheme sequence

\begin{equation*}
G = g_1,g_2,...,g_m
\end{equation*}
with a phoneme sequence

\begin{equation*}
\Phi = \phi_1,\phi_2,...,\phi_n
\end{equation*}
where the sequences are not necessarily of the same length. Data-driven g2p is therefore the problem of finding the phoneme sequence that maximizes the likelihood of the grapheme sequence:

\begin{equation*}
\hat{\Phi} = \underset{\Phi'}{\operatorname{arg\,max}} \Pr(\Phi' \mid G)
\end{equation*}

Data-driven approaches are especially useful for problems in which the rules that govern them are complex and difficult to engineer by hand. g2p for languages with ambiguous orthographies is such a problem. Multilingual g2p, in which the various languages have similar but different and possibly contradictory spelling rules, can be seen as an extreme case of that. Therefore, a data-driven sequence-to-sequence model is a natural choice.

\section{Methods}
\subsection{Encoder--Decoder Models}
In order to find the best phoneme sequence, we use a neural encoder--decoder model with attention \cite{bahdanau2014neural}. The model consists of two main parts:
the \textbf{encoder} compresses each source grapheme sequence $G$ into a fixed-length vector. The \textbf{decoder}, conditioned on this fixed-length vector, generates the output phoneme sequence $\Phi$.

The encoder and decoder are both implemented as recurrent neural networks, which have the advantage of being able to process sequences of arbitrary length and use long histories efficiently. They are trained jointly to minimize cross-entropy on the training data. We had our best results when using a bidirectional encoder, which consists of two separate encoders which process the input in forward and reverse directions. We used long short-term memory units \cite{hochreiter1997long} for both the encoder and decoder. For the attention mechanism, we used the general global attention architecture described by \newcite{luong2015effective}.

We implemented\footnote{\url{https://github.com/bpopeters/mg2p}} all models with OpenNMT \cite{2017opennmt}. Our hyperparameters, which we determined by experimentation, are listed in Table~\ref{table:hyperparameters}.

\begin{table}
\centering
\small
\begin{tabular}{|c|c|}
\hline
\textbf{Enc. \& dec. model type} & LSTM \\
\textbf{Attention} & General \\
\textbf{Enc. \& dec. layers} & 2 \\
\textbf{Hidden layer size} & 150 \\
\textbf{Source embedding size} & 150 \\
\textbf{Target embedding size} & 150 \\
\textbf{Batch size} & 64 \\
\textbf{Optimizer} & SGD \\
\textbf{Learning rate} & 1.0 \\
\textbf{Training epochs} & 13 \\
\hline

\end{tabular}
\caption{Hyperparameters for multilingual g2p models}
\label{table:hyperparameters}
\end{table}

\subsection{Training Multilingual Models}
Presenting pronunciation data in several languages to the network might create problems because different languages have different pronunciation patterns.  For example, the string `real' is pronounced differently in English, German, Spanish, and Portuguese. We solve this problem by prepending each grapheme sequence with an artificial token consisting of the language's ISO 639-3 code enclosed in angle brackets. The English word `real', for example, would be presented to the system as
\begin{center}
\texttt{$<$eng$>$ r e a l}
\end{center}
The artificial token is treated simply as an element of the grapheme sequence. This is similar to the approach taken by \newcite{johnson2016google} in their zero-shot NMT system. However, their source-side artificial tokens identify the target language, whereas ours identify the source language. An alternative approach, used by \newcite{ostling2017continuous}, would be to concatenate a language embedding to the input at each time step. They do not evaluate their approach on grapheme-to-phoneme conversion.

\section{Data}
In order to train a neural g2p system, one needs a large quantity of pronunciation data. A standard dataset for g2p is the Carnegie Mellon Pronouncing Dictionary \cite{lenzo2007cmu}. However, that is a monolingual English resource, so it is unsuitable for our multilingual task. Instead, we use the multilingual pronunciation corpus collected by \newcite{deri2016grapheme} for all experiments. This corpus consists of spelling--pronunciation pairs extracted from Wiktionary. It is already partitioned into training and test sets. 
Corpus statistics are presented in Table~\ref{table:corpus}.

In addition to the raw IPA transcriptions extracted from Wiktionary, the corpus provides an automatically cleaned version of transcriptions. Cleaning is a necessary step because web-scraped data is often noisy and may be transcribed at an inconsistent level of detail. The data cleaning used here attempts to make the transcriptions consistent with the phonemic inventories used in Phoible \cite{phoible}. When a transcription contains a phoneme that is not in its language's inventory in Phoible, that phoneme is replaced by the phoneme with the most similar articulatory features that is in the language's inventory. Sometimes this cleaning algorithm works well: in the German examples in Table~\ref{table:prondata}, the raw German symbols \textipa{/X/} and \textipa{/\c{c}/} are both converted to \textipa{/x/}. This is useful because the \textipa{/X/} in \textit{Ansbach} and the \textipa{/\c{c}/} in \textit{Kaninchen} are instances of the same phoneme, so their phonemic representations should use the same symbol. However, the cleaning algorithm can also have negative effects on the data quality. For example, the phoneme \textipa{/\*r/} is not present in the Phoible inventory for German, but it \emph{is} used in several German transcriptions in the corpus. The cleaning algorithm converts \textipa{/\*r/} to \textipa{/l/} in all German transcriptions, whereas \textipa{/r/} would be a more reasonable guess. The cleaning algorithm also removes most suprasegmentals, even though these are often an important part of a language's phonology. Developing a more sophisticated procedure for cleaning pronunciation data is a direction for future work, but in this paper we use the corpus's provided cleaned transcriptions in order to ease comparison to previous results.

\begin{table}
\centering
\begin{tabular}{|c|cc|}
\hline
Split & Train & Test \\
\hline
\hline
Languages & 311 & 507  \\
Words & 631,828 & 25,894 \\
Scripts & 42 & 45 \\
\hline
\end{tabular}
\caption{Corpus Statistics}
\label{table:corpus}
\end{table}

\begin{table}
\hspace{-0.3em}%
\scalebox{0.8}{
\begin{tabular}{|l|l|l|l|l|}
\hline
{\bf Lang.} & {\bf Script} & {\bf Spelling} & {\bf Cleaned IPA} & {\bf Raw IPA} \\
\hline
deu & Latin & Ansbach &	\textipa{a: n s b a: x} & \textipa{"ansbaX} \\
\hline
deu & Latin & Kaninchen & \textipa{k a: n I n x @ n} & \textipa{ka"ni:n\c{c}@n} \\
\hline
eus & Latin & untxi & \textipa{u \|[n \|[t S I} & \textipa{"un.\t{tS}i} \\
\hline
\end{tabular}}
\caption{Example entries from the Wiktionary training corpus}
\label{table:prondata}
\end{table}

\section{Experiments}
We present experiments with two versions of our sequence-to-sequence model. LangID prepends each training, validation, and test sample with an artificial token identifying the language of the sample. NoLangID omits this token. LangID and NoLangID have identical structure otherwise. To translate the test corpus, we used a beam width of 100. Although this is an unusually wide beam and had negligible performance effects, it was necessary to compute our error metrics.

\subsection{Evaluation}
We use the following three evaluation metrics:
\begin{itemize}
\item Phoneme Error Rate (PER) is the Levenshtein distance between the predicted phoneme sequences and the gold standard phoneme sequences, divided by the length of the gold standard phoneme sequences.
\item Word Error Rate (WER) is the percentage of words in which the predicted phoneme sequence does not exactly match the gold standard phoneme sequence.
\item Word Error Rate 100 (WER 100) is the percentage of words in the test set for which the correct guess is not in the first 100 guesses of the system.
\end{itemize}

In system evaluations,  WER, WER 100, and PER numbers presented for multiple languages are averaged, weighting each language equally \citep[following][]{deri2016grapheme}.

It would be interesting to compute error metrics that incorporate phoneme similarity, such as those proposed by \newcite{hixon2011phonemic}. PER weights all phoneme errors the same, even though some errors are more harmful than others: \textipa{/d/} and \textipa{/k/} are usually contrastive, whereas \textipa{/d/} and \textipa{/\|[d/} almost never are. Such statistics would be especially interesting for evaluating a multilingual system, because different languages often map the same grapheme to phonemes that are only subtly different from each other. However, these statistics have not been widely reported for other g2p systems, so we omit them here.

\subsection{Baseline}
Results on LangID and NoLangID are compared to the system presented by \newcite{deri2016grapheme}, which is identified in our results as wFST. Their results can be divided into two parts:

\begin{itemize}
\item High resource results, computed with wFSTs trained on a combination of Wiktionary pronunciation data and g2p rules extracted from Wikipedia IPA Help pages. They report high resource results for 85 languages.
\item Adapted results, where they apply various mapping strategies in order to adapt high resource models to other languages. The final adapted results they reported include most of the 85 languages with high resource results, as well as the various languages they were able to adapt them for, for a total of 229 languages. This test set omits 23 of the high resource languages that are written in unique scripts or for which language distance metrics could not be computed.
\end{itemize}

\subsection{Training}
We train the LangID and NoLangID versions of our model each on three subsets of the Wiktionary data:

\begin{itemize}
\item LangID-High and NoLangID-High: Trained on data from the 85 languages for which \citet{deri2016grapheme} used non-adapted wFST models.
\item LangID-Adapted and NoLangID-Adapted: Trained on data from any of the 229 languages for which they built adapted models. Because many of these languages had no training data at all, the model is actually only trained on data in 157 languages. As is noted above, the Adapted set omits 23 languages which are in the High test set.
\item LangID-All and NoLangID-All: Trained on data in all 311 languages in the Wiktionary training corpus.
\end{itemize}

In order to ease comparison to Deri and Knight's system, we limited our use of the training corpus to 10,000 words per language. We set aside 10 percent of the data in each language for validation, so the maximum number of training words for any language is 9000 for our systems.

\subsection{Adapted Results}
On the 229 languages for which \newcite{deri2016grapheme} presented their final results, the LangID version of our system outperforms the baseline by a wide margin. The best performance came with the version of our model that was trained on data in all available languages, not just the languages it was tested on. Using a language ID token improves results considerably, but even NoLangID beats the baseline in WER and WER 100. Full results are presented in Table~\ref{table:adapted}.

\begin{table}[h]
\scalebox{0.94}{
\begin{tabular}{lrrr}
\toprule
Model &    WER &  WER 100 &    PER \\
\midrule
wFST             & 88.04  &   69.80  & 48.01 \\
\midrule
LangID-High      & 74.99 &    46.18 & 42.64 \\
LangID-Adapted   & 75.06 &    46.39 & 41.77 \\
LangID-All       & \textbf{74.10} &    \textbf{43.23} & \textbf{37.85} \\
\midrule
NoLangID-High    & 82.14 &    50.17 & 54.05 \\
NoLangID-Adapted & 85.11 &    48.24 & 55.93 \\
NoLangID-All     & 83.65 &    47.13 & 51.87 \\
\bottomrule
\end{tabular}}
\caption{Adapted Results}
\label{table:adapted}
\end{table}

\subsection{High Resource Results}
Having shown that our model exceeds the performance of the wFST-adaptation approach, we next compare it to the baseline models for just high resource languages. The wFST models here are purely monolingual -- they do not use data adaptation because there is sufficient training data for each of them. Full results are presented in Table \ref{table:high}. We omit models trained on the Adapted languages because they were not trained on high resource languages with unique writing systems, such as Georgian and Greek, and consequently performed very poorly on them.

In contrast to the larger-scale Adapted results, in the High Resource experiments none of the sequence-to-sequence approaches equal the performance of the wFST model in WER and PER, although LangID-High does come close. The LangID models do beat wFST in WER 100. A possible explanation is that a monolingual wFST model will never generate phonemes that are not part of the language's inventory. A multilingual model, on the other hand, could potentially generate phonemes from the inventories of any language it has been trained on.

Even if LangID-High does not present a more accurate result, it does present a more compact one: LangID-High is 15.4 MB, while the combined wFST high resource models are 197.5 MB.

\begin{table}[h]
\scalebox{0.94}{
\begin{tabular}{lrrr}
\toprule
Model &    WER &  WER 100 &    PER \\
\midrule
wFST             & \textbf{44.17} &    21.97 & \textbf{14.70} \\
\midrule
LangID-High      & 47.88 &    \textbf{15.50} & 16.89 \\
LangID-All       & 48.76 &    15.78 & 17.35 \\
\midrule
NoLangID-High    & 69.72 &    29.24 & 35.16 \\
NoLangID-All     & 69.82 &    29.27 & 35.47 \\
\bottomrule
\end{tabular}}
\caption{High Resource Results}
\label{table:high}
\end{table}

\subsection{Results on Unseen Languages}
Finally, we report our models' results on unseen languages in Table~\ref{table:unseen}. The unseen languages are any that are present in the test corpus but absent from the training data. Deri and Knight did not report results specifically on these languages. Although the NoLangID models sometimes do better on WER 100, even here the LangID models have a slight advantage in WER and PER. This is somewhat surprising because the LangID models have not learned embeddings for the language ID tokens of unseen languages. Perhaps negative associations are also being learned, driving the model towards predicting more common pronunciations for unseen languages.

\begin{table}[h]
\scalebox{0.94}{
\begin{tabular}{lrrr}
\toprule
Model &   WER &  WER 100 &   PER \\
\midrule
LangID-High      & \textbf{85.94} &    58.10 & \textbf{53.06} \\
LangID-Adapted   & 87.78 &    68.40 & 65.62 \\
LangID-All       & 86.27 &    62.31 & 54.33 \\
\midrule
NoLangID-High    & 88.52 &    58.21 & 62.02 \\
NoLangID-Adapted & 91.27 &    57.61 & 74.07 \\
NoLangID-All     & 89.96 &    \textbf{56.29} & 62.79 \\
\bottomrule
\end{tabular}}
\caption{Results on languages not in the training corpus}
\label{table:unseen}
\end{table}

\section{Discussion}
\subsection{Language ID Tokens}
Adding a language ID token always improves results in cases where an embedding has been learned for that token. The power of these embeddings is demonstrated by what happens when one feeds the same input word to the model with different language tokens, as is seen in Table~\ref{table:tokens}. Impressively, this even works when the source sequence is in the wrong script for the language, as is seen in the entry for Arabic.

\begin{table}[h]
\centering
\begin{tabular}{c|c}
\bf Language & \bf Pronunciation \\
\hline
English & \textipa{d Z u: \ae I s} \\
German & \textipa{j U t s @} \\
Spanish & \textipa{x w i T \|`e} \\
Italian & \textipa{d Z u i t S e} \\
Portuguese & \textipa{Z w i s \~i} \\
Turkish & \textipa{Z U I \|[d Z E} \\
Arabic & \textipa{j u: i s} \\

\end{tabular}
\caption{The word `juice' translated by the LangID-All model with various language ID tokens. The incorrect English pronunciation rhymes with the system's result for `ice'}
\label{table:tokens}
\end{table}

\subsection{Language Embeddings}
Because these language ID tokens are so useful, it would be good if they could be effectively estimated for unseen languages. \newcite{ostling2017continuous} found that the language vectors their models learned correlated well to genetic relationships, so it would be interesting to see if the embeddings our source encoder learned for the language ID tokens showed anything similar. In a few cases they do (the languages closest to German in the vector space are Luxembourgish, Bavarian, and Yiddish, all close relatives). However, for the most part the structure of these vectors is not interpretable. Therefore, it would be difficult to estimate the embedding for an unseen language, or to ``borrow'' the language ID token of a similar language. A more promising way forward is to find a model that uses an externally constructed typological representation of the language.

\subsection{Phoneme Embeddings}
In contrast to the language embeddings, the phoneme embeddings appear to show many regularities (see Table \ref{table:phonemes}). This is a sign that our multilingual model learns similar embeddings for phonemes that are written with the same grapheme in different languages. These phonemes tend to be phonetically similar to each other.

Perhaps the structure of the phoneme embedding space is what leads to our models' very good performance on WER 100. Even when the model's first predicted pronunciation is not correct, it tends to assign more probability mass to guesses that are more similar to the correct one. Applying some sort of filtering or reranking of the system output might therefore lead to better performance.

\begin{table}[h]
\centering
\small
\begin{tabular}{c|c}
\textbf{Phoneme} & \textbf{Closest phonemes} \\
\hline
\textipa{b} & \textipa{p\super h}, \textipa{B}, \textipa{F} \\
\textipa{@} & \textipa{\~a}, \textipa{\u{e}}, \textipa{W} \\
\textipa{t\super h} & \textipa{t:}, \textipa{\:t}, \textipa{\|[t} \\
\textipa{x} & \textipa{X}, \textipa{G}, \textipa{\textcrh} \\
\textipa{y} & \textipa{y:}, \textipa{Y}, \textipa{I} \\
\textipa{\*r} & \textipa{R\super G}, \textipa{\|[r}, \textipa{\;R} \\
\end{tabular}
\caption{Selected phonemes and the most similar phonemes, measured by the cosine similarity of the embeddings learned by the LangID-All model}
\label{table:phonemes}
\end{table}

\subsection{Future Work}
Because the language ID token is so beneficial to performance, it would be very interesting to find ways to extend a similar benefit to unseen languages. One possible way to do so is with tokens that identify something other than the language, such as typological features about the language's phonemic inventory. This could enable better sharing of resources among languages. Such typological knowledge is readily available in databases like Phoible and WALS for a wide variety of languages. It would be interesting to explore if any of these features is a good predictor of a language's orthographic rules.

It would also be interesting to apply the artificial token approach to other problems besides multilingual g2p. One closely related application is monolingual English g2p. Some of the ambiguity of English spelling is due to the wide variety of loanwords in the language, many of which have unassimilated spellings. Knowing the origins of these loanwords could provide a useful hint for figuring out their pronunciations. The etymology of a word could be tagged in an analogous way to how language ID is tagged in multilingual g2p.

\bibliography{references}
\bibliographystyle{emnlp_natbib}

\end{document}